%% file: ijcai25.tex
\documentclass{article}
\makeatletter
\@ifpackageloaded{hyperref}{
  \immediate\write18{bibtex \jobname}
}{}
\makeatother
\usepackage{ijcai25}
\usepackage[utf8]{inputenc} 
\usepackage[T1]{fontenc}    

\usepackage{times}
\usepackage{soul}
\usepackage{url}
\usepackage[hidelinks]{hyperref}
\usepackage[utf8]{inputenc}
\usepackage[small]{caption}
\usepackage{graphicx}
\usepackage{amsmath}
\usepackage{amsthm}
\usepackage{booktabs}
\usepackage{algorithm}
\usepackage{algorithmic}
\usepackage{cite}
\usepackage{amsmath,amssymb,amsfonts}
\usepackage{mathtools}
\usepackage{algorithmic}
\usepackage{booktabs}
\usepackage{multirow}
\usepackage{caption}
\usepackage{graphicx}
\usepackage{textcomp}
\usepackage{float}
\usepackage{xcolor}
\usepackage{amsmath}
\usepackage[switch]{lineno}
\usepackage[numbers]{natbib}
\usepackage{xspace}
\newcommand{\model}{AdaSTI\xspace}
\newcommand{\modelb}{BiS4PI\xspace}
\newcommand{\net}{STC\xspace}
\newcommand{\netb}{NAST\xspace}

\title{\model: Conditional Diffusion Models with Adaptive Dependency Modeling for Spatio-Temporal Imputation}

\author{
    Yubo Yang \and
    Yichen Zhu \and
    Bo Jiang
}

\input{contents/preamble}
\begin{document}

\maketitle

\begin{abstract}
Spatio-temporal data abounds in domain like traffic and environmental monitoring. However, it often suffers from missing values due to sensor malfunctions, transmission failures, etc. Recent years have seen continued efforts to improve spatio-temporal data imputation performance. Recently diffusion models have outperformed other approaches in various tasks, including spatio-temporal imputation, showing competitive performance. Extracting and utilizing spatio-temporal dependencies as conditional information is vital in diffusion-based methods. However, previous methods introduce error accumulation in this process and ignore the variability of the dependencies in the noisy data at different diffusion steps. In this paper, we propose \model (\textbf{Ada}ptive Dependency Model in Diffusion-based \textbf{S}patio-\textbf{T}emporal \textbf{I}mputation), a novel spatio-temporal imputation approach based on conditional diffusion model. Inside \model, we propose a \modelb network based on a bi-directional S4 model for pre-imputation with the imputed result used to extract conditional information by our designed Spatio-Temporal Conditionalizer (\net) network. We also propose a Noise-Aware Spatio-Temporal (\netb) network with a gated attention mechanism to capture the variant dependencies across diffusion steps. Extensive experiments on three real-world datasets show that \model outperforms existing methods in all the settings, with up to 46.4\% reduction in imputation error.

\end{abstract}

\input{contents/introduction}
\input{contents/related_work}

\input{contents/problem}
\input{contents/background}

\input{contents/methodology}
\input{contents/evaluation}

\input{contents/conclusion}

\newpage

\newpage

\input{ijcai25.bbl}
\end{document}

%% file: contents/preamble.tex
\newcommand{\bA}{\mathbf{A}}
\newcommand{\bB}{\mathbf{B}}
\newcommand{\bC}{\mathbf{C}}

\newcommand{\bG}{\mathbf{G}}
\newcommand{\bH}{\mathbf{H}}
\newcommand{\bI}{\mathbf{I}}

\newcommand{\bK}{\mathbf{K}}

\newcommand{\bM}{\mathbf{M}}

\newcommand{\bQ}{\mathbf{Q}}
\newcommand{\bR}{\mathbf{R}}

\newcommand{\bU}{\mathbf{U}}
\newcommand{\bV}{\mathbf{V}}
\newcommand{\bW}{\mathbf{W}}
\newcommand{\bX}{\mathbf{X}}
\newcommand{\bY}{\mathbf{Y}}

\newcommand{\bb}{\mathbf{b}}

\newcommand{\cE}{\mathcal{E}}

\newcommand{\cG}{\mathcal{G}}

\newcommand{\cL}{\mathcal{L}}

\newcommand{\cN}{\mathcal{N}}

\newcommand{\cV}{\mathcal{V}}

\newcommand{\E}{\mathbb{E}}

\newcommand{\R}{\mathbb{R}}

%% file: contents/introduction.tex
\section{Introduction}

Spatio-temporal data captures the intricate dynamics of spatially distributed processes over time. It is prevalent in a wide range of applications, from traffic management to environment monitoring. In real-world scenarios, resulted from sensor malfunctions, transmission errors, etc, there usually exist missing values in spatio-temporal data.



To deal with these missing values, a plethora of methods have emerged for spatio-temporal imputation \citep{cai2015fast,cai2015facets,tong2019netdyna,gong2021spatial,cini2021filling,marisca2022learning,zhu2021networked, liu2023pristi}. They aim to infer the missing values on the condition of observed values and prior knowledge, which is crucial for subsequent data analysis and decision making. 
Among them, benefiting from the powerful capability of generative models in capturing complex dependencies and modeling conditional distributions, Generative Adversarial Network (GAN) \citep{zhu2021networked} and Diffusion \citep{liu2023pristi, chen2024temporal} based methods achieve the state-of-the-art performance.


While diffusion models have already achieved performance on par with GAN-based methods, their potential in spatio-temporal imputation is not yet fully unleashed. A key step in diffusion-based methods is to extract spatio-temporal dependencies and use them as conditions to guide the denoising process for imputation. Existing methods pre-impute the missing entries by linear interpolation, and then use the same spatio-temporal dependencies extracted from the pre-imputation in every denoising step \citep{liu2023pristi, chen2024temporal}. This approach has two drawbacks. First, the pre-imputation by linear interpolation incurs a relatively large error, which then propagates through the entire pipeline. Second, while using the extracted dependencies as the global context prior avoids the difficulty of learning the spatio-temporal dependencies of the noisy data in each denoising step, it also ignores the variability of the latter across different steps, which may compromise the performance.

To overcome the aforementioned drawbacks, we propose \model, a novel spatio-temporal imputation method that models and leverages the step-variant dependencies in conditional diffusion model. \model pre-imputes the missing entries using a trainable network \modelb based on bidirectional S4 layer \citep{gu2021efficiently}, which effectively models the dependencies contained in the observed entries. This provides a more accurate pre-imputation and facilitates the further extraction of spatio-temporal dependencies. To extract step-variant dependencies, we combine both global and local dependencies. We first propose a Spatial-Temporal Conditionalizer (\net) network to extract the global dependencies. Then we propose a Noise-Aware Spatial-Temporal (\netb) network that combines the global dependencies with the local ones in each denoising step by a gated attention mechanism. Such combination accounts for the step-variant nature of  dependencies in diffusion process. In \netb, we also incorporate several effective spatial and temporal modules to enhance its ability to extract spatio-temporal dependencies.

Our contributions are summarized  as follows:
\begin{itemize}
    
    \item We propose {\model}, a novel approach based on conditional diffusion model for spatio-temporal imputation. Our pre-imputation approach facilitates the extraction of spatio-temporal dependencies, and our design of spatio-temporal modules effectively extracts the step-variant dependencies diffusion model's generative process..

    \item We propose a trainable network \modelb to pre-impute incomplete data. The pre-imputation result better reflects the spatio-temporal dependencies 
    and reduces the error accumulation in previous approaches. 
    We also propose a (\netb) network that extracts step-variant dependencies by combining both global and local dependencies with a gated mechanism.
    
    \item We conduct extensive experiments on three real-world datasets under different missing patterns and rates. {\model} outperforms existing methods in all the settings, with up to 46.4\% reduction in imputation error.
\end{itemize}

%% file: contents/related_work.tex
\section{Related Work}
\label{sec:relatedwork}
Given the critical role of complete data in time series analysis, numerous series imputation methods have been proposed. While imputation methods for general time series can be applied to spatio-temporal data,  approaches explicitly exploiting the spatial information can enhance the spatio-temporal imputation performance.

SD-ADMM \citep{meyers2023signal} and WDGTC \citep{li2020tensor} use matrix factorization for time series imputation, which is a light-weighted method based on the low rank hypothesis. DCMF \citep{cai2015fast}, Facet \citep{cai2015facets}, NetDyna \citep{tong2019netdyna} and S-MKKM \citep{gong2021spatial} are matrix factorization methods imputing spatio-temporal data, which take the underlying spatial graph into consideration.

There also exists a large literature addressing time series and spatio-temporal imputation by deep learning. Autoregressive models are suited for modeling series data. BRITS \citep{cao2018brits}, M-RNN \citep{yoon2018estimating} and RDIS \citep{choi2020rdis} leverage RNN \citep{rumelhart1986learning} structure to model the temporal pattern of incomplete series. GRIN \citep{cini2021filling} incorporates MPNN into GRU \citep{chung2014empirical} to capture spatio-temporal correlations. Besides RNN, \citet{shukla2020multi} uses attention mechanism \citep{vaswani2017attention} to learn similarity from incomplete time series and SPIN \citep{marisca2022learning} introduces a sparse spatio-temporal attention mechanism. ImputeFormer \citep{nie2024imputeformer} considers the redundancy in spatio-temporal pattern and imposes a low-rank constraint in transformer. mSSA \citep{agarwal2022multivariate} leverages linear-invariant systems to impute time series.

Among deep learning methods, generative models including VAE, GAN and diffusion models achieve outstanding performance. TimeCIB \citep{choi2023conditional} based on information bottleneck is proposed to impute time series and implemented by VAE. \citet{luo2018multivariate, luo2019e2gan} use GAN framework composed of their proposed GRUI for time series imputation. NETS-ImpGAN \citep{zhu2021networked} is a GAN-based state-of-art spatio-temporal imputation method. It combines GAT \citep{velickovic2018graph} with multi-head self-attention and CNN to capture temporal and spatial correlations. Diffusion model based time series imputation method CSDI \citep{tashiro2021csdi} 
use DiffWave \citep{kong2020diffwave} framework.  CSDI uses an attention mechanism in both temporal and feature dimensions to capture dependencies. 
Another diffusion based state-of-art spatio-temporal imputation methods PriSTI \citep{liu2023pristi} and C\textsuperscript{2}TSD \citep{chen2024temporal} extract spatio-temporal dependencies from data pre-imputed by linear interpolation and use them as conditional information to guide the denoising process. C\textsuperscript{2}TSD further considers trend and seasonal information. However, the utilization of linear interpolation to pre-impute missing entries in diffusion based methods can cause error accumulation, and they also ignore the dependencies across diffusion steps.

%% file: contents/problem.tex
\section{Problem Statement}
\label{sec:problem}

In this paper, we consider spatio-temporal data on a given graph. Let $\cG=(\cV,\cE)$ denote an undirected graph with node set $\cV=\{v_1,v_2,\dots,v_N\}$ and edge set $\cE\subset\cV\times\cV$.  We use  $\bA \in \R^{N \times N}$ to denote the adjacency matrix of $\cG$, whose entries are the edge weights and a larger weight corresponds to a smaller spatial distance between the end nodes. A spatio-temporal data sample over timestamps $\cL=\{1,2,\dots,L\}$ is denoted by $\bX=(X_{n,l})\in\R^{N\times L}$, where $X_{n,l}$ is the value of node $v_n$ at timestamp $l$.

For incomplete data , we use a binary mask $\bM\in\{0,1\}^{N\times L}$ to indicate observed or missing entries of $\bX$: $M_{n,l} = 1$ if $X_{n,l}$ is observed, otherwise $M_{n,l} = 0$. The complementary mask $\overline{\bM}\in\{0,1\}^{N\times L}$ is defined as $\overline{M}_{n,l}=1-M_{n,l}$. We define the observed entries as $\bX^\bM=\{X_{n,l}\mid M_{n,l}=1\}$ and missing entries as $\bX^{\overline\bM}=\{X_{n,l}\mid\overline{M}_{n,l}=1\}$. 
We assume $\bM$ is known, so an incomplete sample is denoted by $(\bX^\bM, \bM)$.  We focus on the case of Missing Completely At Random (MCAR) \citep{little1986statistical}, where the mask $\bM$ is independent of the data $\bX$, i.e., $p(\bM\mid\bX)=p(\bM)$.

Given a sample $(\bX^\bM,\bM)$ of incomplete spatio-temporal data with the underlying graph $\cG$, our task is to impute the missing values $\bX^{\overline{\bM}}$ by sampling from the conditional distribution $p(\bX^{\overline{\bM}}|\bX^{\bM},\bM)$ with the fixed $\cG$. 

%% file: contents/background.tex
\section{Preliminaries}
\label{sec:background}

\subsection{Structured State Space Sequence (S4) Model}
\label{sec:s4}
The Structured State Space Sequence (S4) model is a time series model with the following form, 
\begin{eqnarray}
x'_t = \bA x_t + \bB u_t\text{,} \quad  \quad y_t = \bC x_t. 
\label{SSM}
\end{eqnarray}
where $u_t$ is the input series, $y_t$ is the output, and $x_t$ is the hidden state. S4 learns the parameters $\bA$, $\bB$ and $\bC$ by initializing $\bA$ to a special class of  matrices called the HiPPO matrices \citep{gu2020hippo}. These matrices assigns uniform weights to all history values. In other commonly used time series models, inputs closer to the current step have a greater impact on the current state. As a result, the absence of inputs from the previous time step significantly affects the subsequent states. In contrast, initialization by HiPPO in S4 helps the model to mitigate the impact of missing data from previous time steps.


\citet{gu2021efficiently} discretize and expand Eqn.~\eqref{SSM} into the following discrete time system:
\begin{eqnarray}
y_t = \overline{\bC}\overline{\bA}^{t}\overline{\bB} u_0 + \overline{\bC}\overline{\bA}^{t-1}\overline{\bB} u_t + ... + \overline{\bC}\overline{\bB} u_t,
\label{SSM_conv}
\end{eqnarray}
where $\overline{\bA}, \overline{\bB}$ and $\overline{\bC}$ are  matrices derived from $\bA, \bB$ and $\bC$. S4 computes Eqn.~\eqref{SSM_conv} using a convolution form for efficiency: $y = \overline{\bK}*u$, where $\overline{\bK}= (\overline{\bC}\overline{\bA}^{i}\overline{\bB})_{i\in [t]}$ is the convolution kernel. S4 model shows great performance in sequence modeling tasks.

\subsection{Denoising Diffusion Probabilistic Models} 
\label{sec:ddpm}
 DDPM is trained through a forward noise injection process and generates the distribution $p_\theta(\bX)$ to approximate the true distribution $p(\bX)$ through a reverse denoising process. The forward process continually adds random Gaussian noise to  $\bX_0=\bX$ for $T$ times:
\begin{eqnarray}
\begin{aligned}
    p(\bX_{1:T}\mid\bX_0) \coloneqq \prod_{t=1}^{T}p(\bX_t|\bX_{t-1})\quad \text{and} \\ \quad p(\bX_t|\bX_{t-1})\coloneqq\cN(\sqrt{1-\beta_t}\bX_{t-1},\beta_t\bI),
\end{aligned}
\end{eqnarray}
where $\beta_t\in(0,1)$ controls the level of added noise at step $t$. $X_t$ can be represented by a closed-form: $\bX_t = \sqrt{1-\beta_t}\bX_{t-1} + \sqrt{\beta_t} \epsilon$, $\epsilon \sim \cN(0, \bI)$. The forward process also allows sampling of $\bX_t$ at an arbitrary step $t$: using the notation $\alpha_t \coloneqq 1-\beta_t$ and $\overline{\alpha}_t \coloneqq\prod_{s=1}^{t}\alpha_s$, we have $p(\bX_t|\bX_0) = \cN(\bX_t;\sqrt{\overline{\alpha}_t}X_0,(1-\overline{\alpha}_t)\bI)$. Similarly, the closed-form is $\bX_t = \sqrt{\overline{\alpha}_t}\bX_0 + \sqrt{1-\overline{\alpha}_t}\epsilon$. Therefore, given an arbitrary noise level $t$,  we can train the model to directly predict the total noise added at step $t$ without training through each individual noise adding step.
 
The reverse process aims to recover the original $\bX_0$ from $\bX_T$ to generate samples. It is represented as:
\begin{eqnarray}
p_\theta(\bX_{0:T})\coloneqq p(\bX_T)\prod_{t=1}^T p_\theta(\bX_{t-1}\mid\bX_t){,}
\label{eq:sample1}
\end{eqnarray}
\citet{feller2015retracted} proves that when $\beta_t$ is small, $p(\bX_{t-1}|\bX_t)$ is also a Gaussian distribution. It is estimated as $p_\theta(\bX_{t-1}|\bX_t)$ through parameterizing its mean and covariance by employing a neural network in DDPM:
\begin{eqnarray}
p_\theta(\bX_{t-1}|\bX_t)\coloneqq \cN(\bX_{t-1};\mu_\theta(\bX_t,t), \sigma_\theta(\bX_t,t)\bI) \text{,}
\label{eq:sample2}
\end{eqnarray}
The mean and covariance matrix of $p_\theta(\bX_{t-1}|\bX_t)$ are
\begin{eqnarray}
\begin{aligned}
&\mu_\theta(\bX_t,t)=\frac{1}{\alpha_t}\left(\bX_t-\frac{\beta_t}{\sqrt{1-\alpha_t}}\epsilon_\theta(\bX_t,t)\right),\\ &\sigma_\theta(\bX_t,t)=\hat\beta_t^{\frac{1}{2}}\text{.}
\end{aligned}
\end{eqnarray}
where $\epsilon_\theta$ is a trainable function to predict the noise added to $\bX_t$. The hyperparameter $\hat\beta_t = \frac{1-\alpha_{t-1}}{1-\alpha_t}\beta_t$.

The neural network can be trained by optimizing:
\begin{eqnarray}
\min\limits_\theta\ell(\theta)\coloneqq \E_{\bX_0\sim p(\bX_0),\epsilon\sim\cN(\mathbf{0},\bI),t}\|\epsilon-\epsilon_\theta(\bX_t,t)\|_2^2\text{.}
\end{eqnarray}

%% file: contents/methodology.tex
\section{Methodology}
\begin{figure}[t] 
\center{\includegraphics[width=0.52\textwidth]  {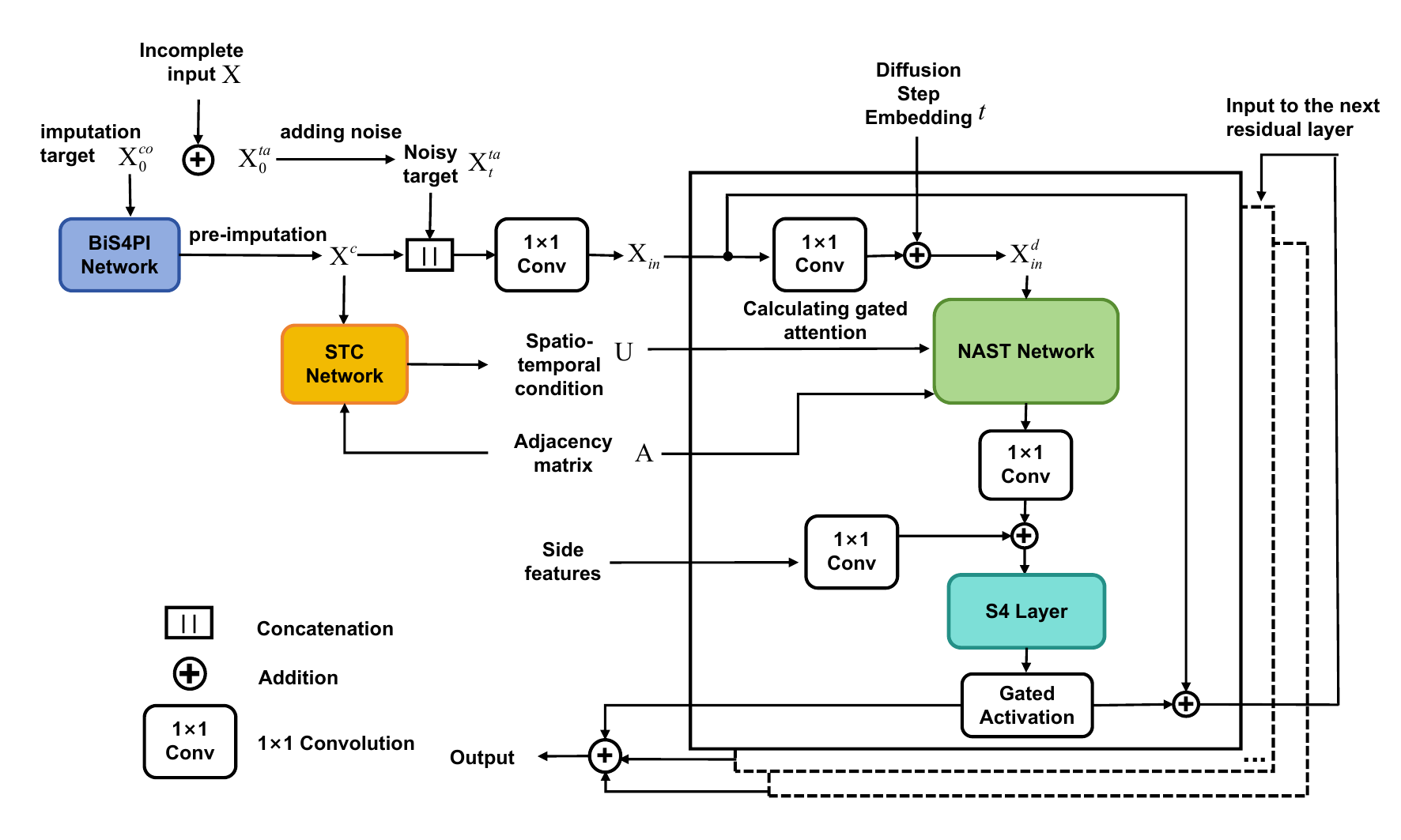}}
\captionsetup{font=small}
\vspace{-20pt}
\caption{Architecture of {\model}}
\vspace{-10pt}
\label{fig:architecture}
\end{figure}

\subsection{Conditional DDPM For Spatio-temporal Imputation}
\label{fram}
The way we use conditional diffusion model for spatio-temporal imputation is inspired by CSDI \citep{tashiro2021csdi} for time series imputation. Given conditional data $(\bX^{\bM^{co}},\bM^{co})$ and the graph $\cG$, we aim to impute the target data $(\bX^{\bM^{ta}},\bM^{ta})$. $\bX^{\bM^{co}}$ is the observed entries and $\bX^{\bM^{ta}}$ is the ground truth of the missing entries. We introduce two new notations $\bM^{co}$ and $\bM^{ta}$ instead of using $\bM$ and $\overline\bM$, the reason for which will become clear later. Different from DDPM, the goal is to learn the conditional distribution $p(\bX^{\bM^{ta}}\mid\bX^c,\bM^{ta},\bM^{co},\bU)$, where $\bX^c$ is the pre-imputed data 
and $\bU$ is the spatio-temporal conditional information extracted from $\bX^c$.

The forward process is very similar to DDPM, since adding noise is independent of the conditional information. The difference is that we only need to add noise to the imputation target entries $\bX^{\bM^{ta}}$ instead of the entire sample. 

The reverse process aims to recover the original target data $\bX^{\bM^{ta}}_0$ from the noisy $\bX^{\bM^{ta}}_T$: 
\begin{eqnarray}
\begin{aligned}
&p_\theta(\bX^{\bM^{ta}}_{0:T}\mid\bX^c,\bM^{ta},\bM^{co},\bU) \coloneqq \\ &p(\bX^{\bM^{ta}}_T) \prod_{t=1}^{T}p_\theta(\bX^{\bM^{ta}}_{t-1}\mid\bX^{\bM^{ta}}_t,\bX^c,\bM^{ta},\bM^{co},\bU)\text{,}
\end{aligned}
\end{eqnarray}

In addition to the target data $(\bX^{\bM^{ta}}_t,\bM^{ta})$ and mask $\bM^{co}$ of the next step, we also exploit the pre-imputation result $\bX^c$, conditional spatio-temporal information $\bU$ and graph $\cG$ to model each reverse step. Specifically, the mean and covariance are parameterized as
\begin{eqnarray}
\begin{aligned}
&\mu_\theta(\bX^{\bM^{ta}}_t,t,\bX^c,\bM^{ta},\bM^{co},\bU,\cG) \coloneqq \frac{1}{\alpha_t} (\bX^{\bM^{ta}}_t-  \\&
\frac{\beta_t}{\sqrt{1-\alpha_t}}\epsilon_\theta(\bX^{\bM^{ta}}_t,t,\bX^c,\bM^{ta},\bM^{co},\bU,\cG))
\end{aligned}
\label{mu_theta}
\end{eqnarray}
and
\begin{eqnarray}
\sigma_\theta(\bX^{\bM^{ta}}_t,t,\bX^c,\bM^{ta},\bM^{co},\bU,\cG)=\hat\beta_t^{\frac{1}{2}}\text{.}
\label{sigma_theta}
\end{eqnarray}
The function $\epsilon_\theta$ is implemented with a trainable neural network, which will be introduced in Section \ref{sec:arc}. The reverse process can be trained by optimizing the loss function:
\begin{eqnarray}
\begin{aligned}
\min\limits_\theta\ell(\theta)\coloneqq & \E_{\bX_0\sim p(\bX_0),\epsilon\sim\cN(\mathbf{0},\bI),t} \|\epsilon-\epsilon_\theta\|_2^2,
\end{aligned}
\label{loss1}
\end{eqnarray}
As a input to $\epsilon_\theta(\bX^{\bM^{ta}}_t,t,\bX^c,\bM^{ta},\bM^{co},\bU,\cG))$, $\bX^{\bM^{ta}}_t$ is the target to be denoised in the reverse process. It is obtained by adding noise to $\bX^{\bM^{ta}}_0$ in the forward diffusion process $\bX^{\bM^{ta}}_t = \sqrt{\overline{\alpha}_t}\bX^{\bM^{ta}}_0 + \sqrt{1-\overline{\alpha}_t}\epsilon$. This results in $\bX^{\bM^{ta}}_0$ having to be treated as known in the training process, which is unavailable in the incomplete data scenario. As a result, we cannot use the missing data as target data, i.e., set $\bM^{ta}=\overline\bM$.

To overcome this issue, we select part of the observed data as imputation target and treat them as missing, of which we do have the ground truth, and regard the rest of the observed data as conditional data in the training process. Our strategy of selection of target data follows CSDI \citep{tashiro2021csdi}. $\bM^{ta}$ and $\bM^{co}$ satisfy $\bM^{ta} \cup \bM^{co} = \bM$.
In this way, our diffusion model can be trained with only observed data. 

In each iteration, after obtaining a training sample $\bX$, we first choose imputation target $\bX^{\bM^{ta}}_0$ from the observed parts and get $(\bM^{ta}, \bM^{co})$. Then, we pre-impute the incomplete data to obtain $\bX^c$ and extract spatio-temporal conditions $\bU$ from it. Afterward, we randomly sample Gaussian noise $\epsilon$, diffuse it with a randomly chosen number $t$ to add noise on $\bX^{\bM^{ta}}_0$, obtaining $\bX^{\bM^{ta}}_t$, and  at last input them into the network to obtain the predicted noise $\epsilon_\theta$. Consequently, we can use the training objective Eqn.~\eqref{loss1} to train our noise prediction network. 

In the evaluation imputation process, we set $\bM^{ta}=\overline\bM$ and $\bM^{co}=\bM$. The imputed data $\bX^{\bM^{ta}}_0$ can be obtained by recursively going through the reverse process. We fill the missing entries with random Gaussian noise and use $\epsilon_\theta$ to denoise with Eqn.~\eqref{mu_theta} and Eqn.~\eqref{sigma_theta}. We repeat this process for $k$ times and take the median as the final imputation. For each entry in $\bX^{\bM^{ta}}$, the amount of conditional data $\bX^{\bM^{co}}$ is different between training and test since we split out the target data from the observed data during training. To mitigate such difference, we select only few entries as training target data. 

\subsection{Architecture of Denoising Function}
\label{sec:arc}
In this section, we provide a detailed introduction of the architecture of \model, which is shown in Fig.\ref{fig:architecture}. The incomplete input is first pre-imputed by the \modelb and the \net network extracts spatio-temporal dependencies $\bU$ from the pre-imputation $\bX^c$. Linear interpolation in previous approaches \citep{liu2023pristi, chen2024temporal} introduce relatively large pre-imputation error, which can mislead the extraction and cause error accumulation. S4 model assigns uniform weights to all history values. The absence of a portion of the data prior to the current step does not significantly affect the current state , which makes it applicable in the incomplete data scenario. The bidirectional design makes \modelb take the entire sequence into account during pre-imputation, better reflecting the pattern of the sequence.

Then we concatenate $\bX^c$ and the noisy target $\bX_t^{ta}$ together and process it by 1D-convolution layer to get $\bX_{in}$ as the input to the noise prediction network $\epsilon_\theta$. It adopts the residual structure in DiffWave \citep{kong2020diffwave}, which previous approaches \citep{tashiro2021csdi, liu2023pristi, chen2024temporal, hu2023towards} have shown to be highly effective in diffusion models for series data. The key component of each residual layer is our designed \netb network. Previous methods use the dependencies extracted from the pre-imputation as invariant global condition but ignore the variability of the dependencies of the noisy data in each denoising step. Inside \netb, we take both the global spatio-temporal dependencies and those varying in each denoising step into consideration, which enhance imputation performance. The output of the noise prediction network is the estimated noise added to the imputation target at the noise adding step $t$. We provide the details of each component of \model in the following. 

\textbf{Pre-imputation by BiS4PI Network.} 
To address the impact of missing data on extracting dependencies, we propose a trainable neural network based on the bidirectional S4 model to pre-impute the incomplete spatio-temporal data. 

\begin{figure}[htb] 
\vspace{-10pt}
\center{\includegraphics[width=0.45\textwidth]  {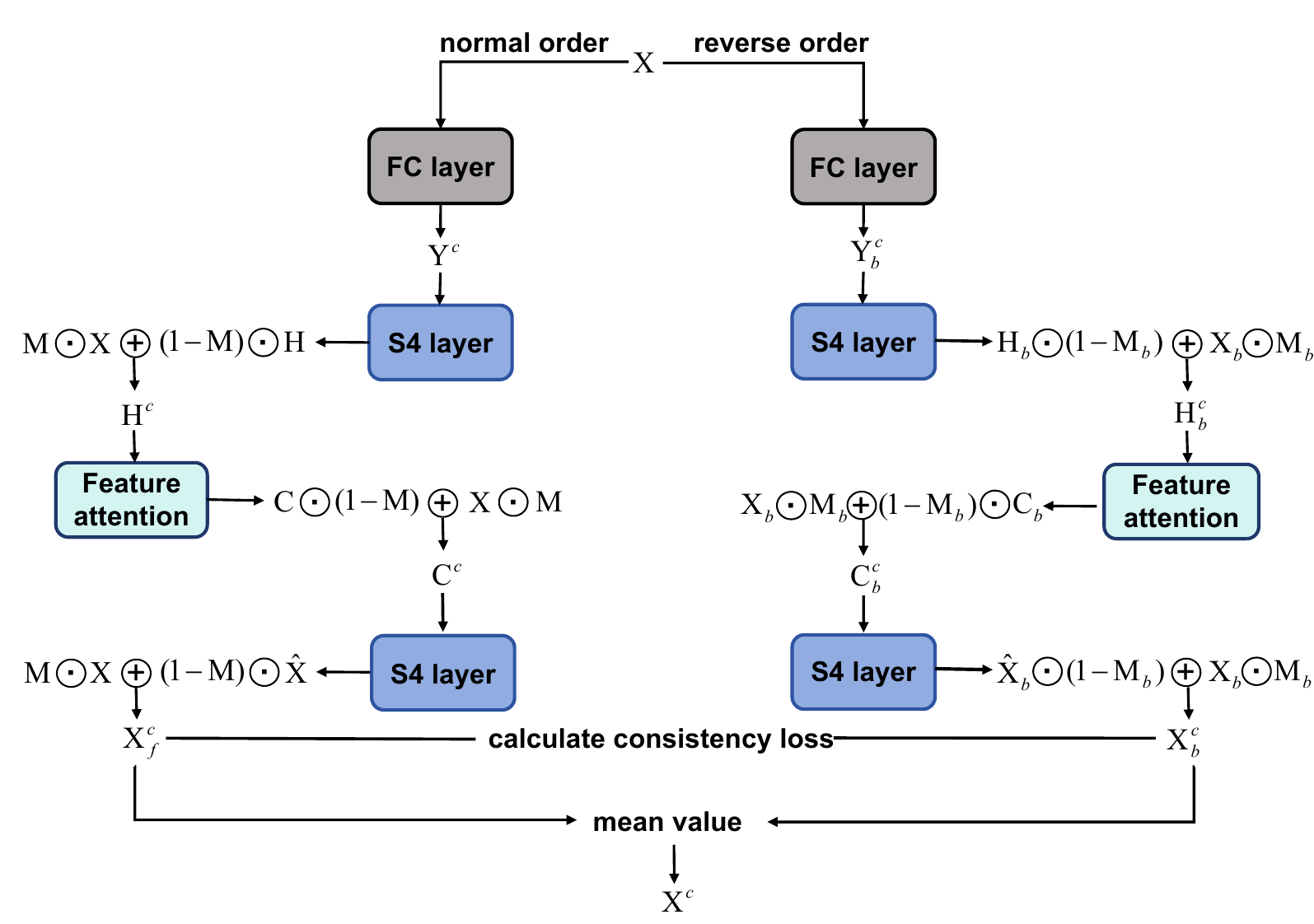}}
\caption{Architecture of \modelb.}
\label{fig:bis4}
\vspace{-10pt}
\end{figure}

The architecture of \modelb is shown in Fig. \ref{fig:bis4}. In each network, in order to diminish the impact of inputting a sample with missing entries into the S4 model, we first utilize a linear layer to fill in the missing parts of the input sample $\bX$.
\begin{eqnarray}
\bY^c = \bW_x\bX + \bb_x \text{.}
\end{eqnarray}
Input $\bX \in \R^{L \times N}$, where $L$ is the length of the input sequence and $N$ is the number of nodes. $\bY^c$ has the same shape with $\bX$ and is fed into the first S4 layer with the convolution form introduced in \ref{sec:s4}: $\bH = \overline{\bK}*\bY^c$. Then, we replace missing parts of $\bX$ by $\bH$:
\begin{eqnarray}
\bH^c = \bX\odot\bM + \bH\odot(1-\bM)\text{.}
\end{eqnarray}
where $\odot$ is the element-wise multiplication. So far, we have primarily used temporal patterns to pre-impute incomplete data by S4 layer. To better reflect the dependencies between nodes, we employ the attention mechanism on the feature dimension where each feature represents a node. We calculate attention along $\bH^c$ feature dimension with a transformer encoder layer: $\bC = \mathrm{Attn}_{\mathrm{Feature}}(\bH^c)$. The coarse imputation $\bC$ at timestamp $l$ considers both the historical observations and the ones at the current timestamp. Then we proceed to replace missing entries in $\bX$ by $\bC$ and obtain $\bC^c$, and input it into the second S4 layer to derive the network's output $\hat{\bX}$:
\begin{align}
\bC^c &= \bX\odot\bM + \bC\odot(1-\bM), \quad \hat{\bX} = \mathrm{S4}(\bC^c) \text{.}
\end{align}
In the final step, we replace the output corresponding to the observed entries in $\hat{\bX}$ with $\bX$: $\bX^c = \bX\odot\bM + \hat{\bX}\odot(1-\bM)$. We deploy two identical networks to take the whole sequence into consideration, which operate on the input sequence in the forward and reverse direction from the temporal perspective respectively. Their results are $\bX^c_{f}$ and $\bX^c_{b}$, and we use the mean value as the final output $\bX^c$.

\textbf{\net Network.} The \net network is designed for extracting spatio-temporal dependencies from $\bX^c$ as conditional information to guide the reverse process of the diffusion model for generating samples. 
We incorporate effective spatial and temporal modules inside \net to extract spatio-temporal dependencies. It takes $\bX^c$ and the adjacency matrix $\bA$ as inputs and outputs the spatio-temporal conditional information $\bU$:
\begin{align}
\bU_{in} &= \mathrm{Conv(\bX^c)}  \text{,}
\label{STDE1}\\
\bY_{tem} &= \mathrm{Norm}(\mathrm{Attn}_{tem}(\bU_{in}) + \bU_{in}) \text{,}
\label{STDE2}\\
\bY_{GCN} &= \mathrm{Norm}(\mathrm{GCN}(\bY_{tem}, \bA)+\bU_{in})  \text{,}
\label{STDE3}\\
\bY_{spa} &= \mathrm{Norm}(\mathrm{Attn}_{spa}(\bY_{tem})+\bU_{in})  \text{,}
\label{STDE4}\\
\bY_{sum} &= \bU_{in} + \bY_{tem} + \bY_{GCN} + \bY_{spa}  \text{,}
\label{STDE5}\\
\bU &= \mathrm{Norm}(\mathrm{MLP(\bY_{sum})} + \bY_{sum})  \text{.}
\label{STDE6}
\end{align}
where $\mathrm{Conv(\cdot)}$ is the convolution layer, $\mathrm{Attn}_{tem}(\cdot)$ is the self-attention along the temporal dimension, $\mathrm{GCN(\cdot)}$ is the Graph Convolution Network \citep{wu2019graph} and $\mathrm{Attn}_{spa}(\cdot)$ is the self-attention along the feature dimension.

\begin{figure}[htb] 
\vspace{-12pt}
\center{\includegraphics[width=0.4\textwidth]  {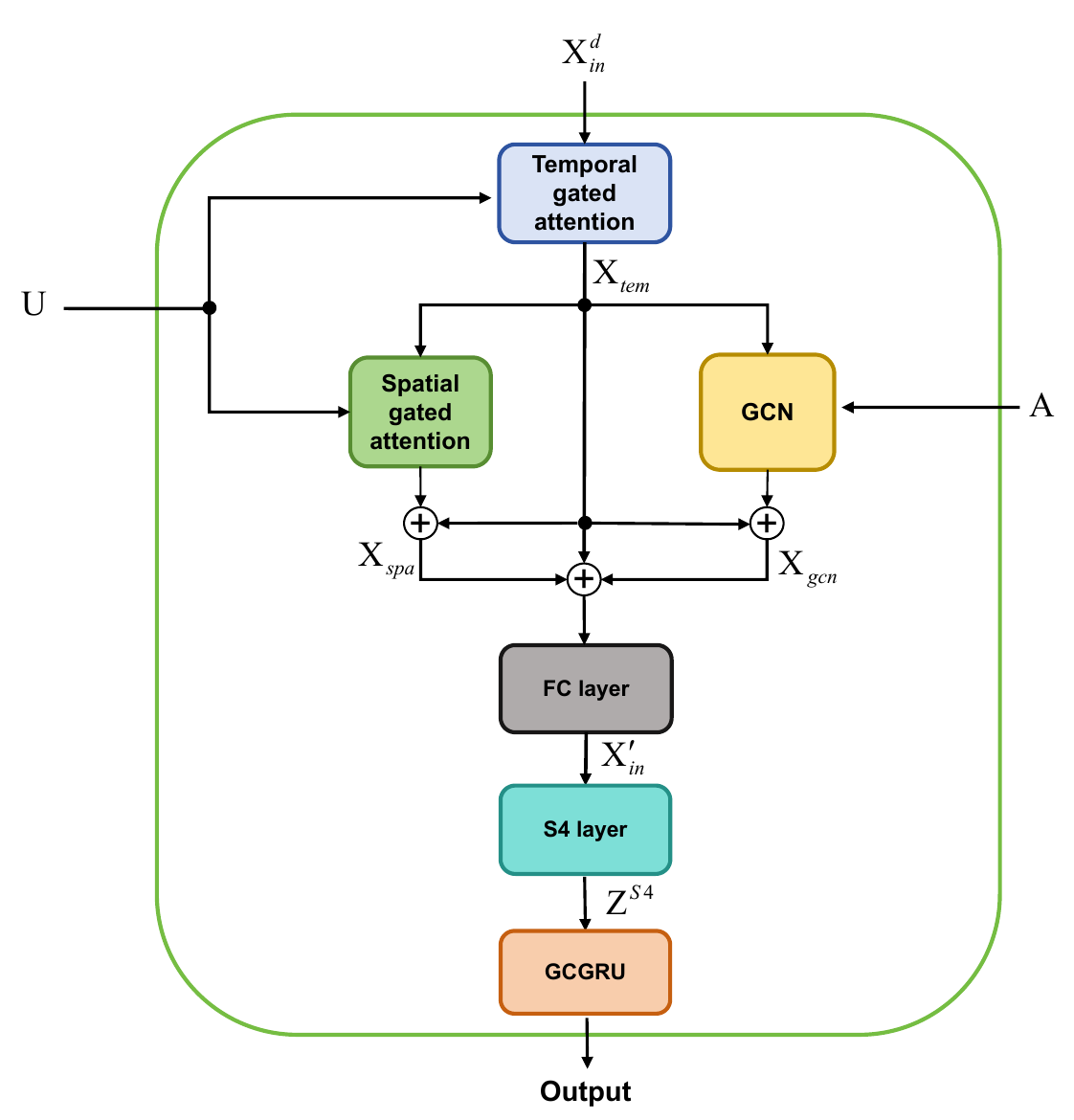}}
\vspace{-10pt}
\caption{Architecture of {\netb} network}
\label{fig:nast}
\end{figure}

\textbf{\netb Network.} The \netb network is the main components in each residual layer of $\epsilon_{\theta}$, whose architecture is shown in Fig. \ref{fig:nast}. 
\netb has the gated-controlled attention mechanism incorporated with GCN architectur. The outputs of the attention layers are obtained by taking a trainable weighted sum of the outputs from the cross-attention layer and the self-attention layer. 

For a normal self-attention layer $\mathrm{Attn(\cdot)}$ with the input $\bX_{in}$, its output is formulated as: $\mathrm{Attn(\bQ, \bK, \bV)} = \mathrm{softmax}(\frac{\bQ\bK^T}{\sqrt{d_1}})\bV$, where $d_1$ is the hidden dimension and $\bQ, \bK, \bV$ are query, key, value matrix calculated by:
\begin{eqnarray}
\bQ = \bX_{in}\bW^Q \text{,} \quad \bK = \bX_{in}\bW^K  \text{,} \quad \bV = \bX_{in}\bW^V \text{.}
\end{eqnarray}
where $\bW^Q, \bW^K, \bW^V$ are learnable parameter matrices. The difference in cross-attention layer is that the $\bQ$ and $\bK$ matrices are calculated by conditional information $\bU$:
\begin{eqnarray}
\bQ = \bU\bW_c^Q \text{,} \quad \bK = \bU\bW_c^K \text{,} \quad \bV =\bX_{in}\bW_c^V \text{.}
\end{eqnarray}
Note that our cross-attention is different from USTD \citep{hu2023towards}, whose value and key matrix are obtained by the historical representations. We denote the output of the self-attention layer as $\bR_{self}$ and the output of the cross-attention layer as $\bR_{cross}$. The output $\bR$ of the gated attention layer is:
\begin{eqnarray}
\bR = \bG\odot\bR_{self} + (1-\bG) \odot \bR_{cross} \text{,}\\
\bG = \sigma(\bW_{g1}\bR_{self} + \bW_{g2}\bR_{cross} + \bb_g)\text{.}
\end{eqnarray}
where $\bW_{g1}, \bW_{g2}, \bb_g$ are parameters. The gated attention layer is applied along the temporal and feature dimension respectively. The forward process of the gated-attention incorporated with GCN architecture is formulated as:
\begin{align}
\bX_{tem} & = \mathrm{GateAttn}_{tem}(\bX_{in}^d,\bU)  \text{,}\label{STDU1} \\
\bX_{gcn} & =  \mathrm{Norm}(\mathrm{GCN}(\bX_{tem}, \bA) + \bX_{tem}) \label{STDU2} \text{,}
\\
\bX_{spa} & =  \mathrm{Norm}(\mathrm{GateAttn}_{spa}(\bX_{tem},\bU) + \bX_{tem})\text{,}\label{STDU3} \\
\bX^{'}_{in} \hspace{5pt} & = \mathrm{Norm}(\mathrm{MLP}(\bX_{gcn} + \bX_{spa})) \label{STDU4} \text{.}
\end{align}
To enhance the ability to capture spatial and temporal dependency, we incorporate additional S4 layer and GCGRU \citet{cini2021filling} in \net. 

\begin{table*}[htb]
\centering
\scalebox{0.75}{\begin{tabular}{@{}c rr @{\hspace{12pt}} rr rr@{\hspace{18pt}} rr@{\hspace{18pt}} rr}
\toprule
\multirow{3}{*}{Method} & \multicolumn{2}{c}{Metro (25\% random)} & \multicolumn{2}{c}{Metro (25\% block)} & \multicolumn{2}{c}{Electricity (25\% random)} & \multicolumn{2}{c}{Electricity (25\% block)} & \multicolumn{2}{c}{AQ36}  \\
\cmidrule(lr){2-3} \cmidrule(lr){4-5} \cmidrule(lr){6-7} \cmidrule(lr){8-9} \cmidrule(lr){10-11} &  MAE & RMSE & \hspace{1em} MAE & RMSE &  \hspace{1em} MAE & RMSE & \hspace{1.1em} MAE & RMSE & \hspace{.5em} MAE & RMSE \\
\midrule
Mean & 149.52 & 268.54 & 149.53 & 268.13 & 4.64 & 11.33 & 4.58 & 11.29 & 24.17 & 49.02 \\
TLI & 27.49 & 51.87 & 27.99 & 50.33 & 1.16 & 3.68 & 1.19 & 3.85 & 13.93 & 30.83 \\
\midrule
S-MKKN & 22.66& 41.77& 22.65 &43.60 & 1.08 & 3.08 & 1.13 & 3.49 & 14.47 & 31.81 \\
WDGTC & 21.26 & 41.00 & 22.06 & 42.05 & 1.05 & 3.28 & 1.06 & 3.19 & 12.24 & 26.42 \\
DCMF & 20.94 & 37.50 & 21.60 & 39.30 & 0.91 & 2.31 & 0.94 & 2.62 & 12.33 & 25.80 \\
\midrule
BRITS & 23.26 & 41.96 & 23.47 & 43.76 & 0.99 & 2.92 & 0.96 & 2.85 & 13.28 & 27.97 \\
mSSA & 23.65 & 46.75 & 23.91 & 45.27 & 0.97 & 2.71 & 0.99 & 2.68 & 12.82 & 26.80 \\
GRIN & 21.02 & 37.77 & 21.92 & 40.12 & 0.63 & 1.28 & 0.65 & 1.34 & 11.45 & 23.36 \\
Imputeformer & 18.90 & 30.91 &  21.77  & 34.53 & 0.53 &  1.31 & 0.65 & 1.49 & 12.54 & 26.53 \\
\midrule
TimeCIB & 22.22 & 37.68 & 22.30 & 37.82 & 0.71 & 1.38 & 0.82 & 1.66 & 11.06 & 22.73 \\
CSDI & 21.66 & 41.48 & 21.17 & 41.06 & 0.73 & 1.62 & 0.70 & 1.60 & 10.23 & 21.33 \\
PriSTI & 18.97 & 31.43 & 19.41 & \underline{31.28} & \underline{0.56} & \underline{1.10} & \underline{0.54}& \underline{1.11} & \underline{9.42} & \underline{18.19} \\
NETS-ImpGAN & \underline{17.40} & \underline{30.79} & \underline{18.18} & 32.87 & \underline{0.56} & 1.17 & \underline{0.54} & 1.23 & 9.61 & 18.99 \\
\midrule
{\model} & \textbf{16.99} & \textbf{27.75} & \textbf{17.38} & \textbf{28.23} & \textbf{0.30} & \textbf{0.79} & \textbf{0.32} & \textbf{1.03} & \textbf{9.03} & \textbf{18.08} \\
\bottomrule
\end{tabular}}
\captionsetup{font=footnotesize}
\caption{Imputation performance on Metro (25\% random, 25\% block), Electricity (25\% random, 25\% block) and AQ36. \textbf{Bold} and \underline{underline} indicate the best and second best in each row.}

\label{tab:1}
\vspace{-5pt}
\end{table*}

\begin{table*}[htb]

\centering
\scalebox{0.75}{\begin{tabular}{@{}c r r r r r r r r r r r r}
\toprule
 \multirow{2}{*}{Method} &\multicolumn{2}{c}{2\%} &  \multicolumn{2}{c}{6\%} &\multicolumn{2}{c}{10\%} & \multicolumn{2}{c}{25\%} & \multicolumn{2}{c}{50\%} & \multicolumn{2}{c}{75\%}\\
 \cmidrule(lr){2-3} \cmidrule(lr){4-5} \cmidrule(lr){6-7}\cmidrule(lr){8-9}\cmidrule(lr){10-11}\cmidrule(lr){12-13} ~ &MAE &RMSE &MAE &RMSE &MAE &RMSE  &MAE &RMSE &MAE &RMSE &MAE &RMSE\\

\midrule
Mean &147.71 &265.61 & 147.61&265.79  & 149.13 & 268.45&149.52&268.54&147.83&265.68 &146.67&262.81 \\ 
TLI &24.82&47.17 & 25.91&48.56  & 25.79 & 48.31&27.49& 51.87&30.77&58.60&43.67&81.85\\
\midrule
S-MKKN & 22.37&40.61  & 22.05 & 39.97&21.83&39.90 &22.66 &41.77 &25.70& 49.50& 31.14& 58.56\\
WDGTC &21.71 &41.33 & 21.20&40.46  & 20.39 &  40.39&21.26&41.00&24.44&45.05&29.35& 55.89 \\
DCMF  &19.23 &33.71 & 19.89&35.11  & 20.18 & 37.17&20.94&37.50&24.75&47.58&30.48&57.43 \\
\midrule
BRITS &22.82 &41.08 & 22.55&39.93& 22.34 & 40.56&23.26&41.96&26.59&50.87&31.65&60.01 \\
mSSA &23.48 & 45.55 & 22.78& 44.98&23.04&44.98&23.65 & 46.75&27.15 &53.48 & 32.71& 63.44\\
GRIN & 19.04&34.05  &20.02& 34.65 & 20.53 & 37.28&21.02&37.77&24.46&47.82&30.49&58.02\\
Imputeformer & 18.14 &  28.69 &  18.37 &  28.92 & 18.41 & 29.42 & 18.90 &30.91 & 19.77 &32.14 &21.95 & 37.38\\
\midrule
TimeCIB & 20.81 & 34.03 & 21.06 & 33.70 &  22.01 & 37.65 & 22.22 & 37.68 & 23.64 &39.46 & 26.65 & 46.78 \\
CSDI &21.36 &40.30  & 21.12&40.24 & 21.04 & 40.13&21.66&41.48&24.74&45.74&29.82&56.31\\
Pri-STI & 17.26 &\underline{28.43} & 17.91 & 29.90 & 18.51 &30.62 & 18.97& 31.43& 20.68 & \underline{33.85} &25.28 &  \underline{43.72}\\
NETS-ImpGAN &\underline{16.37} & 28.85  &\underline{16.74} &  \underline{29.63} & \underline{17.11} & \underline{30.30} & \underline{17.40} & \underline{30.79}& \underline{20.60}&39.23& \underline{24.77} &47.34\\
\midrule
{\model} &\textbf{16.08} & \textbf{25.23} & \textbf{16.18} &\textbf{25.74}  &\textbf{16.37}&\textbf{26.17}  &\textbf{16.99} & \textbf{27.75}&\textbf{18.58} &\textbf{30.54}&\textbf{21.44}&\textbf{36.38} \\ 
\bottomrule
\end{tabular} }
\captionsetup{font=footnotesize}
\caption{Imputation performance on Metro dataset under 2\%-75\% missing rate in the random missing pattern. \textbf{Bold} and \underline{underline} indicate the best and second best in each row.}

\label{table:metro_random}
\vspace{-15pt}
\end{table*}
\begin{table*}[htb]

\centering
\scalebox{0.75}{\begin{tabular}{@{}c r r r r r r r r r r r r}
\toprule
 \multirow{2}{*}{Method} &\multicolumn{2}{c}{2\%} &  \multicolumn{2}{c}{6\%} & \multicolumn{2}{c}{10\%} & \multicolumn{2}{c}{25\%} & \multicolumn{2}{c}{50\%} & \multicolumn{2}{c}{75\%}\\
 \cmidrule(lr){2-3} \cmidrule(lr){4-5} \cmidrule(lr){6-7}\cmidrule(lr){8-9}\cmidrule(lr){10-11}\cmidrule(lr){12-13} ~ &MAE &RMSE &MAE &RMSE &MAE &RMSE  &MAE &RMSE &MAE &RMSE &MAE &RMSE\\

\midrule
Mean & 147.31 & 266.46   &147.98 & 264.57 &149.54&268.95&149.53&268.13 & 147.62&264.43&146.51&262.15 \\ 
TLI &24.69 & 46.17  &25.56 & 47.86  &25.33 & 47.81 & 27.99 & 50.33 & 31.10 & 59.52 & 44.04& 83.04\\
\midrule
S-MKKN &21.51& 39.66&22.32&39.94&21.96& 39.78&22.65&43.60&26.10&48.52&32.04& 59.80 \\
WDGTC &21.17&41.12&21.94&40.38&20.49&42.18&22.06& 42.05&24.62& 43.23&30.35&57.02 \\
DCMF & 18.45 & 33.15  & 20.64 &  35.40 & 20.61 & 36.29&21.60 & 39.30 & 25.14&47.69 & 31.31 &58.66\\
\midrule
BRITS & 22.85 &40.51&22.88&40.06&21.91&42.01&23.47&43.76&27.03&49.85&31.52&60.77\\
mSSA &23.85&44.18&22.29&44.11&22.93&46.56&23.91&45.27&26.84&55.23&32.41& 62.93 \\
GRIN &18.03&34.21&20.29&34.88&21.07&39.15&21.92& 40.12&24.53&49.16&31.38& 58.03\\
Imputeformer & 20.37 & 30.19 & 20.78 & 33.72 & 21.03 & 34.49  & 21.77& 34.53 & 21.99 & 36.94 & 26.05 & 49.15 \\
\midrule
TimeCIB & 21.45  & 35.63 & 22.30 & 35.67 & 22.31 & 36.62 & 22.53 & 38.15 & 24.05 & 40.75 & 30.45 & 55.39 \\
CSDI &21.67&40.29&21.45&38.26&21.38&38.58&21.17&41.06&24.49&45.78&30.29&56.41\\
Pri-STI & 17.38 &\underline{28.15} & 17.81 & \underline{29.79} & 19.06 & \underline{31.02}& 19.41 & \underline{32.28}& 21.40&\underline{35.39} &28.39 & 53.06\\
NETS-ImpGAN &\underline{17.13}&30.41& \underline{17.40}& 31.73& \underline{18.04}&31.58& \underline{18.18}& 32.87&\underline{21.20}&41.19&\underline{25.36}& \underline{49.13}\\
\midrule
{\model} & \textbf{16.17} & \textbf{25.25}  &\textbf{16.31} &\textbf{25.26} &\textbf{16.75} &\textbf{25.99} &\textbf{17.38} & \textbf{28.23}&\textbf{19.48} &\textbf{32.97}&\textbf{24.46}&\textbf{45.39} \\
\bottomrule
\end{tabular} }
\captionsetup{font=footnotesize}
\caption{Imputation performance on Metro dataset under 2\%-75\% missing rate in the block missing pattern. \textbf{Bold} and \underline{underline} indicate the best and second best in each row.}
\label{table:metro_block}
\end{table*}

\textbf{Training of The Whole Framework.} The training loss of of our model is the sum of three components: the loss function of the conditional diffusion model Eqn.~ \eqref{loss1}, the reconstruction loss and the consistency loss for supervising the pre-imputation of \modelb.

The reconstruction loss is calculated between the observed entries of the input data $\bX$ and the corresponding entries of the three stages of reconstruction in \modelb: $\bH^c$, $\bC^c$ and $\hat{\bX}$. Specifically, we utilize the Mean Absolute Error (MAE) loss. The total reconstruction loss consists of three parts: $\ell(\bX\odot\bM, \bH^c\odot\bM)$, $\ell(\bX\odot\bM, \bC^c\odot\bM)$ and $\ell(\bX\odot\bM, \hat{\bX}\odot\bM)$. For the sake of convenience, we denote them as $\ell_1$, $\ell_2$ and $\ell_3$. The total reconstruction loss is the sum of them:
\begin{eqnarray}
\ell_{rec} = \ell_1 + \ell_2 + \ell_3\text{.}
\end{eqnarray}
The above refers to the reconstruction loss along the forward temporal order and we just need to reverse the order of the input $\bX$ in case of the backward temporal order. We denote the total reconstruction loss of forward and backward direction as $\ell_{rec}^{f}$ and $\ell_{rec}^{b}$ respectively.

Beyond the computation of reconstruction errors of the observed entries, it is important to consider the quality of pre-imputation for the missing entries. However, the absence of ground truth precludes a direct computation of the pre-imputation loss. Therefore, we introduce an additional loss function that penalizes inconsistencies between the forward and reverse pre-imputation of the missing entries like \citep{cao2018brits}:
\begin{eqnarray}
\ell_{cons} = \mathrm{MAE}(\bX^c_f\odot(1-\bM), \bX^c_b\odot(1-\bM)) \text{.}
\end{eqnarray}

The total objective function of the framework is denoted as:
\begin{eqnarray}
\begin{aligned}
\min\limits_\theta\ell_{total}(\theta)\coloneqq \E_{\bX_0\sim p(\bX_0),\epsilon\sim\cN(\mathbf{0},\bI),t} \|\epsilon- \epsilon_\theta(\bX^{\bM^{ta}}_t,t, 
\\
\bX^c, \bM^{ta},\bM^{co},\bU,\cG)\|_2^2 + \lambda(\ell_{rec}^f + \ell_{rec}^b + \ell_{cons}) \text{.}
\label{loss2}
\end{aligned}
\end{eqnarray}
where $\lambda$ is a hyper-parameter.

%% file: contents/evaluation.tex
\section{Evaluation}
\label{sec:evaluation}

\subsection{Experimental Setup}
\label{sec:exeset}
\textbf{Datasets.} We evaluate the performance of {\model} on the following traffic, energy supply and environment datasets:
\begin{itemize}
    \item \textbf{Hangzhou Metro Passenger Flow (Metro)} \citep{tianchi2019metro} is metro passenger outbound flow data collected from 81 metro stations in Hangzhou. It is recorded every 10 minutes from January 1st to January 25th in 2019.
    \item \textbf{Electricity Consumption (Electricity)} \citep{Irish2016electricity} records electricity consumption in Ireland within every 30 minutes from 2009 to 2010 from 427 smart meters. 
    \item \textbf{AQ36} \citep{yi2016st} collects hourly PM2.5 pollution data from 36 sensors located in Beijing, monitored from May 2014 to April 2015.
\end{itemize}
The adjacency matrices of the datasets are calculate by thresholded Gaussian kernel \citep{shuman2013emerging}. AQ36 dataset is naturally incomplete with a missing rate of 13.24\% and the missing pattern is not random. We use the same strategy as \citet{tashiro2021csdi} in choosing imputation targets on AQ36 in the training. Metro and Electricity are complete datasets. We generate incomplete datasets for them according to random and block missing patterns. 1) \textbf{random}: All nodes at all timestamps are independently missing; 2) \textbf{block}: The block missing pattern we use is the same as the pattern in NETS-ImpGAN \citep{zhu2021networked}. Multiple blocks are missing at random positions. In each block, values on $N_v$ adjacent nodes at randomly selected $N_t$ consecutive timestamps are missing.

\textbf{Baselines.} We compare \model with simple methods: mean, temporal linear interpolation (TLI) and state-of-art methods: matrix factorization: S-MKKM \citep{gong2021spatial}, WDGTC \citep{li2020tensor}, DCMF \cite{cai2015fast}; generative models: TimeCIB \citep{choi2023conditional},  CSDI \citep{tashiro2021csdi},  Pri-STI \citep{liu2023pristi}, NETS-ImpGAN \citep{zhu2021networked}; other methods: BRITS \citep{cao2018brits}, mSSA \citep{agarwal2022multivariate}, GRIN \citep{cini2021filling}, ImputeFormer \citep{nie2024imputeformer}.

\textbf{Metrics.} We use mean absolute error (MAE) and root mean square error (RMSE) as the evaluation metrics. As a probabilistic imputation method based on DDPM, we randomly sample Gaussian noise 100 times to denoise and take the median of denoising results as the imputation result.


\subsection{Imputation Performance}
\label{sec:exeper} 
We evaluate the performance of \model under different missing rates and patterns. Table \ref{tab:1} shows the results under both missing patterns, 25\% missing rate on Metro, Electricity and naturally incomplete AQ36. \model consistently outperforms all the baselines in terms of both MAE and RMSE on all three datasets. \model outperforms diffusion based method PriSTI, which also considers both spatial and temporal dependencies on all datasets and settings. This comparison shows that our specially designed networks can extract and utilize the spatio-temporal dependencies as conditional information better.

We also study the impact of missing rates. Tables \ref{table:metro_random} and \ref{table:metro_block} show the results on Metro under different missing rates for the random and block missing patterns, respectively. \model outperforms all the baselines under all the missing rates. On average of 2\%-75\% missing rate, we reduce RMSE by 12.4\% and 11.5\% in random and block missing patterns with respect to the second best achieved by NETS-ImpGAN or PriSTI. MAE are reduced by 12.4\% and 11.5\% in the two missing patterns. As the missing rate increases, the performance of \model degrades much slower than the baselines, especially in RMSE. This indicates that the benefits of \model become more significant as the missing rate increases and shows \model generalization ability in different missing rates.

\subsection{Ablation Study}
\label{sec:exeabl}
In this section, we study the efficacy of key components in {\model} and the impact of the model complexity on performance. 

For the two key components, we remove them from \model respectively: (1) Remove the pre-imputation network \modelb ({\model}-w/o-\modelb); (2) Change the gated attention mechanism to cross-attention ({\model}-w/o-gated attention). Table \ref{tab:abl} shows the results on the naturally incomplete AQ36 dataset and Metro dataset (25\% random). The full model {\model} 
achieves the best performance, which shows the efficacy of the components in extracting, utilizing conditional information and capturing spatial-temporal dependencies.

\begin{table}[h]
\centering
\scalebox{0.75}{
\begin{tabular}{c r r@{\hspace{12pt}} r r}
\toprule

 \multirow{2}{*}{Method} &\multicolumn{2}{c}{Metro (25\% random)} & \multicolumn{2}{c}{AQ36}  \\
 \cmidrule(lr){2-3} \cmidrule(lr){4-5} & \hspace{.2em} MAE &   RMSE & \hspace{.5em} MAE & RMSE\\
\midrule

{\model}-w/o-\modelb &17.33 & 28.79& 9.22&18.88 \\
{\model}-w/o-gated attention&17.38 &28.40 & 9.17& 18.65\\
\midrule
{\model} & \textbf{16.99} &\textbf{27.75} & \textbf{9.03} &\textbf{18.08} \\
\bottomrule
\end{tabular} }
\captionsetup{font=small}
\caption{\model vs. ablated variants.}
\vspace{-10pt}
\label{tab:abl}
\end{table}

Model complexity, often measured by the number of trainable parameters, can be critical for the performance. We calculate the number of trainable parameters in \model: 10876221 and diffuion based state-of-art method PriSTI: 739590. The difference primarily arises from the attention layers within the residual blocks. Firstly, our gated attention mechanism consists of self-attention and cross-attention layers, whereas PriSTI employs only cross-attention. Moreover, the size of the projection layers following the attention output is different between \model and PriSTI. Table \ref{tab:ab2} shows the performance of \model and PriSTI under different parameters settings.
\begin{table}[h]
    \centering
    \scalebox{0.75}{
\begin{tabular}{c c r r}
\toprule

 \multirow{2}{*}{Method} & \multirow{2}{*}{Parameters} &\multicolumn{2}{c}{Metro (25\% random)}\\
 \cmidrule(lr){3-4} & & \hspace{.2em}  MAE &  RMSE \\
\midrule

 PriSTI-64$\times$64 & 739590  & 18.97 & 31.43\\
\model-64$\times$64 & 2686269  & 17.18   & 28.10 \\
 \model-64$\times$2048 &  10876221 & \textbf{16.99} & \textbf{27.75} \\
\bottomrule
\end{tabular} }
    \caption{Impact of model complexity}
    \label{tab:ab2}
\vspace{-10pt}
\end{table}
The $a \times b$ following the model name indicates the size of the projection layers after attention. To ensure a rigorous comparison, we reduce \model's projection size from 64$\times$2048 to 64$\times$64, causing only slight performance drop. With the same projection size, \model still outperforms PriSTI by 9.4\%. This improvement is not due to an increase of total model parameters but is achieved through our gated attention mechanism.


%% file: contents/conclusion.tex
\section{Conclusion}
\label{sec:conclusion}
In this paper, we propose \model, which is a state space model enhanced spatio-temporal data imputation approach based on the conditional diffusion model. To better extract spatio-temporal dependencies, we employ \modelb to perform pre-imputation. \modelb facilitates extraction of conditional information and mitigates the error accumulation. We also design \net to extract spatio-temporal conditional information from the pre-imputation and \netb to leverage these dependencies within the conditional diffusion model. Inside \netb, we design a gated attention mechanism to comprehensively consider the extracted spatio-temporal dependencies, as well as those varying with the variant distribution of denoising targets across different denoising steps.
We conduct extensive experiments on three real-world datasets under different missing patterns and missing rates. The results show that {\model} outperforms existing methods in all the settings.